\documentclass[letterpaper, 10 pt, conference]{ieeeconf}  
\usepackage{times}
\usepackage{soul}
\usepackage{url}
\usepackage[hidelinks]{hyperref}
\usepackage[small]{caption}
\usepackage{graphicx}

\let\labelindent\relax
\usepackage{amsmath}
\usepackage{booktabs}
\usepackage{algorithm}
\usepackage{algorithmic}
\usepackage{times}
\usepackage{helvet}
\usepackage{courier}
\usepackage{amsthm}
\usepackage{graphicx}
\usepackage{booktabs}
\usepackage{color}
\usepackage{tabularx}
\usepackage{hyperref}
\usepackage{caption}
\usepackage{subcaption}
\usepackage{booktabs}
\usepackage{multirow}
\usepackage{float}
\usepackage{longtable}
\theoremstyle{definition}
\usepackage{nameref}
\newtheorem{definition}{Definition}[section]
\usepackage{enumitem}

\IEEEoverridecommandlockouts                              

\title{\LARGE \bf
Learning a Safety Verifiable Adaptive Cruise Controller from Human Driving Data}

\author{Qin Lin$^{1}$, Sicco Verwer$^{2}$, and John Dolan$^{1}$
\thanks{$^{1}$Qin Lin and John Dolan are with the Robotics Institute, Carnegie Mellon University, Pittsburgh, PA 15213, USA {\tt\small qinlin,jdolan@andrew.cmu.edu}}%
\thanks{$^{2}$Sicco Verwer is with the Department of Intelligent Systems, Delft University of Technology, 2600 AA Delft, the Netherlands {\tt\small s.e.verwer@tudelft.nl}}%
}
\begin{document}

\maketitle
\thispagestyle{empty}
\pagestyle{empty}

\begin{abstract}
Imitation learning provides a way to automatically construct a controller by mimicking human behavior from data. For safety-critical systems such as autonomous vehicles, it can be problematic to use controllers learned from data because they cannot be guaranteed to be collision-free. Recently, a method has been proposed for learning a multi-mode hybrid automaton cruise controller (MOHA). Besides being accurate, the logical nature of this model makes it suitable for formal verification. In this paper, we demonstrate this capability using the SpaceEx hybrid model checker as follows. After learning, we translate the automaton model into constraints and equations required by SpaceEx. We then verify that a pure MOHA controller is not collision-free. By adding a safety state based on headway in time, a rule that human drivers should follow anyway, we do obtain a provably safe cruise control. Moreover, the safe controller remains more human-like than existing cruise controllers. 
\end{abstract}

\section{INTRODUCTION}
Adaptive cruise control (ACC) systems assist drivers to maintain safety spacing from leading vehicles and ease the workload of frequent acceleration and deceleration operations. A key drawback of existing ACCs is the inconsistency between systems and human driving habits, since the control algorithm of an ACC is based on mathematical optimization of safety and comfort rather than mimicking actual driving behaviors \cite{hiraoka2005}. An alternative approach is imitation learning, which mimics human control strategies in order to obtain behavior that is similar to the driving trajectories of human drivers. As a representative work, convolutional neural networks (CNNs) have been successfully applied to map raw pixels from a single front-facing camera directly to steering commands \cite{bojarski2016}. Being a safety-critical system, however, it is very important to know whether an imitation learning cruise controller is safe to use, i.e., whether it can cause collisions or not. In \cite{tian2018} such a study is performed. They use simulations to test the safety properties of controllers based on deep neural networks. We argue, however, that since unexpected situations will at some point occur in practice, testing these properties in simulations is insufficient.

Recently, an imitation-learning-based model named \textbf{m}ulti-m\textbf{o}de \textbf{h}ybrid \textbf{a}utomaton (MOHA) has been proposed to mimic car-following behaviors of human drivers \cite{lin2018}. This model includes both \emph{discrete observations} and \emph{continuous output actions}. The observations are obtained by discretizing signal values such as speed and distance to the leading vehicle. The output controls the acceleration pedal of the following vehicle.

In this paper, we demonstrate that the logical nature of MOHA controller allows it to be formally verified using the SpaceEx hybrid system model checker \cite{frehse2011}. This was recently achieved for a simplified traditional (not learned) ACC in \cite{mishra2016}. 

The main idea of our work is to use SpaceEx to verify whether collisions are avoided by MOHA when given a non-deterministic leading vehicle. The leading vehicle is only constrained by vehicle dynamics, e.g., it can produce any trajectory falling within physically possible speed and acceleration ranges. To achieve this, we develop a transformation \emph{MO2SX} from the discrete observations, which trigger state transitions in MOHA to a set of linear inequalities that can be used by SpaceEx. The transformation is exact without approximation error. In addition, we enhance MOHA to include actions for any possible future action, including those that never occurred in the training data but might be tested by the model checker.

We perform experiments in a variety of traffic for both highway and urban driving scenarios. The experiments demonstrate that purely learning a MOHA controller from data is unsafe, e.g., that it can collide in extreme cases. We then add a \emph{safety state} to MOHA (a common addition to ACC systems). Essentially, the controller is forced to push the brake if the time needed to reach the current position of the leading vehicle drops below, for instance, two seconds suggested in the highway driving scenarios. We show that:
\begin{itemize}
    \item MOHA controller with safety state is guaranteed to be collision-free.
    \item MOHA is more safe, more accurate, and more human-like than existing controllers and neural networks.
\end{itemize}
These results demonstrate clear advantages of using explainable models based on logic (such as MOHA) over black-box models (such as neural nets) for imitation learning. Instead of trusting an AI-based controller based on simulations, our work demonstrates the possibility to verify with certainty whether an AI-based controller is safe. 

The remainder of this work is organized as follows. Related work is reviewed in Section \ref{sec:S2}. MOHA and its learning algorithm are introduced in Section \ref{sec:S3}. The model transformation for hybrid model checker is explained in Section \ref{sec:S4}. The experimental results are discussed in Section \ref{sec:S5}. Our conclusions are in Section \ref{sec:S6}. 

\section{Related work}\label{sec:S2}
Verifying the safety of hybrid models is known to be undecidable except for severely restricted models such as timed automata and initialized rectangular automata \cite{alur1995}. There exist three categories of techniques/tools that address relaxed versions of this problem.

The first category is \emph{deductive verification}, which combines user interaction with an automated
theorem prover in a proof search utilizing differential logics \cite{loos2013}. KeYmaera is the dominant tool in this category, and has been used for safety verification of vehicle-to-vehicle (V2V) communication in ACCs \cite{loos2013}.

The second category is \emph{symbolic reachability analysis}, which includes tools such as HyTech \cite{henzinger1997} for linear hybrid automata, d/dt \cite{asarin2002d}, PHAVER \cite{frehse2005}, SpaceEx \cite{frehse2011} for piecewise linear affine dynamics, and Flow* \cite{chen2013} for non-linear dynamics. In these techniques, symbolic reachability algorithms iteratively explore reachable states starting from the initial states. There is no termination guarantee because the algorithm may reach more and more states without being able to conclude that the system is safe. In practice, setting a maximum number of states, a fixed-point reaching criterion, or a maximum running time is used to force 
termination. In very related work, a highly simplified ACC with constant acceleration and deceleration in an open loop control system is verified using symbolic reachability analysis in SpaceEx \cite{mishra2016}.

The third category is called \emph{abstraction}. The main idea is obtaining an abstraction of 
coarse dynamics over the original model. Proving the safety of the abstract model then is a sufficient condition for proving the corresponding properties in the original model \cite{henzinger1998}. The drawback is that it can be difficult to avoid an oversimplification.

In this work, we use symbolic reachability analysis using SpaceEx, similar to the work of \cite{mishra2016} but using a complex model that has been learned from data.

Also related are recent works on generating test cases for neural networks. Deep neural networks are a popular method for learning dynamics such as those in ACCs. DeepXplore \cite{pei2017} and DeepTest \cite{tian2018} propose white-box and gray-box methods for automated generation of test cases and discovering the corner cases from deep neural networks (DNN). However, they focus more on software logic testing using a coverage criterion. This type of testing is incomplete and does not perform a full reachability analysis. Recently, researchers make progress on verifying safety of deep neural network-based controllers using reachability analysis techniques.
Feedforward neural networks with piecewise linear ReLUs activation functions are demonstrated to be verifiable \cite{xiang2019reachable,tran2019parallelizable}. More complex deep neural net with sigmoid activation functions are also verified \cite{ivanov2019verisig}. However, the architecture of these neural network is limited to feedfarward one. In addition, verifying these NN-based controllers is essentially in a \emph{non-dynamical} fashion, i.e., statically verifying input-output properties at each time step. Our work is dealing with a \emph{dynamical} verification problem: given a sequence of environment observations and control actions, we verify whether the safety property is guaranteed. Indeed, combining reachable states from multiple steps can achieve a sequential verification results, however, we will suffer from large over-approximation error and high computation cost problems.

\section{MOHA: An hybrid automaton model}\label{sec:S3}
\theoremstyle{definition}
\begin{definition}{Hybrid automaton:}
A hybrid automaton $H$ is a tuple $<\textbf{Loc}, \textbf{Edge},
\mathbf{\Sigma}, \mathbf{X}, \textbf{Init}, \textbf{Inv}, \textbf{Flow}, \textbf{Jump}>$ where:
\begin{description}[style=unboxed,leftmargin=0cm]
\item[$\bullet$ Loc] is a finite set $\{l_1, l_2, \cdots, l_m\}$ of (control) locations that represent control modes of the hybrid system, which are essentially discrete states in a finite state automaton.
\item[$\bullet$ $\mathbf{\Sigma}$] is a finite set of events.
\item[$\bullet$ Edge] $\subseteq \textbf{Loc} \times \mathbf{\Sigma} \times \textbf{Loc}$ is a finite set of labeled edges representing discrete changes of control modes in the hybrid system. Those changes are labeled by events from $\mathbf{\Sigma}$.
\item[$\bullet$ X] is a finite set $\{x_1, x_2,\cdots, x_n\}$ of $n$-dimension real-valued variables. For example, in a standard ACC system, the variables at least include the position of the leading and following vehicles $x_l$ and $x_f$, and their speeds $v_l$ and $v_f$. $\dot{X}$ is for the first-order differential of variables $\{\dot{x}_1, \dot{x}_2, \cdots, \dot{x}_m\}$ inside a location. The primed variables $\{x'_1, x'_2, \cdots, x'_n\}$ are used to represent updates of variables from one control mode to another, called an assignment.
\item[$\bullet$ Init(l)] is a predicate for the valuation of free variables from $X$ when the hybrid system starts from location $l$.
\item[$\bullet$ Inv(l)] is a predicate whose free variables are from X. It constrains the possible valuations for those variables when the control of the hybrid system is in location $l$. A commonly used convex predicate is a finite conjunction of linear inequalities, e.g. $x_1 \geq 3 \wedge 3x_2 \leq x_3+5/2.$, which represents a polytope consisting of multiple half-spaces.
\item[$\bullet$ Flow(l)] is a predicate whose free variables
are from $X \cup \dot{X}$. It defines a continuous system evolution for when the control mode is in location $l$ using a differential equation (usually ordinary differential equation, ODE).
\item[$\bullet$ Jump] is a function that assigns to each labeled edge a predicate whose free variables are from $X \cup \dot{X}$. Jump($e$) states when the discrete change modeled by the event $e$ is possible and what the variable updates are when the hybrid system makes this discrete change.
\end{description}
\end{definition}

\begin{figure}[t]
\centering
\includegraphics[width=0.48\textwidth]{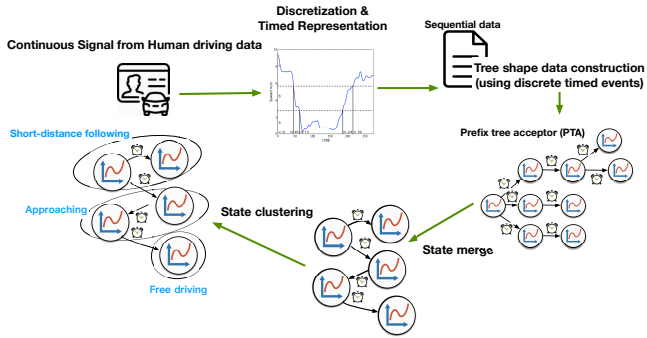}
\caption{Flowchart illustrating MOHA learning. The discretization on a one-dimension signal is just for a simple demonstration. The original signal is multidimensional. Also, MOHA shows more than 3 modes in car-following behaviors \protect\cite{lin2018}. \label{fig:moha_flowchart}}
\end{figure}

MOHA is a novel model for learning car-following behaviors using a hybrid automaton recently proposed in \cite{lin2018}. The main idea of learning MOHA for continuous time series data is illustrated in the flowchart shown in Fig. \ref{fig:moha_flowchart}.

First, continuous variables from time series are discretized into sequences of symbolic events. Each sequence is a complete car-following trajectory from a pair consisting of a leading vehicle and a following vehicle.
The time gap between two consecutive events is encoded in order to represent time-varing behaviors, e.g., moderate/harsh braking. In this way, we obtain timed strings $\{(e_1^i, t_1^i), \cdots (e_j^i, t_j^i), \cdots, (e_n^i, t_n^i)\}$ from the $i$-th trajectory, where $t_j^i$ is the time difference between discrete events $e_j^i$ and $e_{j-1}^i$.

Second, as a model for the discrete dynamics, a timed automaton is learned using the RTI+ real-time identification algorithm \cite{verwer2010}. 
The original continuous values used to obtain the corresponding discretized values in the timed string are stored in every state.

Third, states are partitioned based on a state subsequence clustering, i.e., several states in a subsequence cluster are grouped into one block in the automaton. These blocks form the different control modes of the ACC system.

Last, numeric data reached in distinct modes are used to 
identify the parameters of differential equations in these modes 
using differential evolution algorithms (DEA).

The environmental input in MOHA is 3-dimensional, i.e., the relative speed, the relative distance, and the following vehicle's speed.
Changes to these variables may trigger 
discrete state and control mode transitions.
After entering a new mode, the controller uses 
the corresponding differential equation 
to generate continuous acceleration/deceleration 
output.

These equations are 
linear Helly models \cite{hel1959}. The acceleration in Helly's model is a linear function combining the relative speed ($\Delta v = v_l-v_f$) and the relative distance between the headway ($\Delta x = x_l-x_f$) and the desired headway, which is defined by :

\begin{equation}\label{equ:helly1}
\dot{v_f}(t) = C_1 \cdot \Delta v(t) + C_2 \cdot \left( \Delta x(t) - D(t) \right)
\end{equation}
and
\begin{equation}\label{equ:helly2}
D(t) = \alpha + \beta \cdot v_f(t)
\end{equation}
where $C_1$, $C_2$, $\alpha$, $\beta$, are the constant parameters that need to be calibrated. The desired headway is a function of the speed of the following vehicle and a safety distance, where $\alpha$, $\beta$ and are the corresponding weightings for those variables. Note that, compared with the original Helly model, we neglect time delays because the SpaceEx model checker does not support tracking long historical variables, so all computations are on-the-fly.

\section{Hybrid model checker}\label{sec:S4}
Hybrid model verification based on reachability computation is similar in spirit to \emph{numerical simulation}, which produces all possible trajectories one by one to check whether the system behaves properly. 
The obvious drawback is the fact that all possible trajectories are non-enumerable, though it has been a popular ``verification'' approach in several ACC design works \cite{eyisi2013}. The reachability algorithm explores the state space in a breadth-first manner, that is, at each time step all the states reachable by all possible one-step inputs from states reachable in the previous step. Though the computation is costly, it provides more confidence in the correctness of the system than a small number of individual simulated trajectories. In the hybrid verification problem, an over-approximation is used for the set of reachable states, and a conventional symbolic state reachability algorithm is used. By  checking whether forbidden states such as collisions are reachable, the model can be guaranteed to be safe.

\subsection{SpaceEx}
SpaceEx is a powerful and popular tool for safety verification of hybrid systems. It supports hybrid systems with linear piecewise affine and non-deterministic dynamics, i.e., $\mathbf{\dot{X}=AX+b}$, where $\mathbf{b}$ is non-deterministic turbulence. SpaceEx consists of three main components: \emph{Model editor} is a graphical editor for creating models of complex hybrid systems. \emph{Analysis core} is a command line program that takes a model file in \emph{.xml} format, and a configuration file \emph{.cfg} that specifies the initial states. \emph{Web interface} is a graphical user interface with which one can specify initial states and other analysis parameters, run the analysis core, and visualize the output.
\subsection{Translator}
\begin{figure}[t]
\centering
\includegraphics[width=0.5\textwidth]{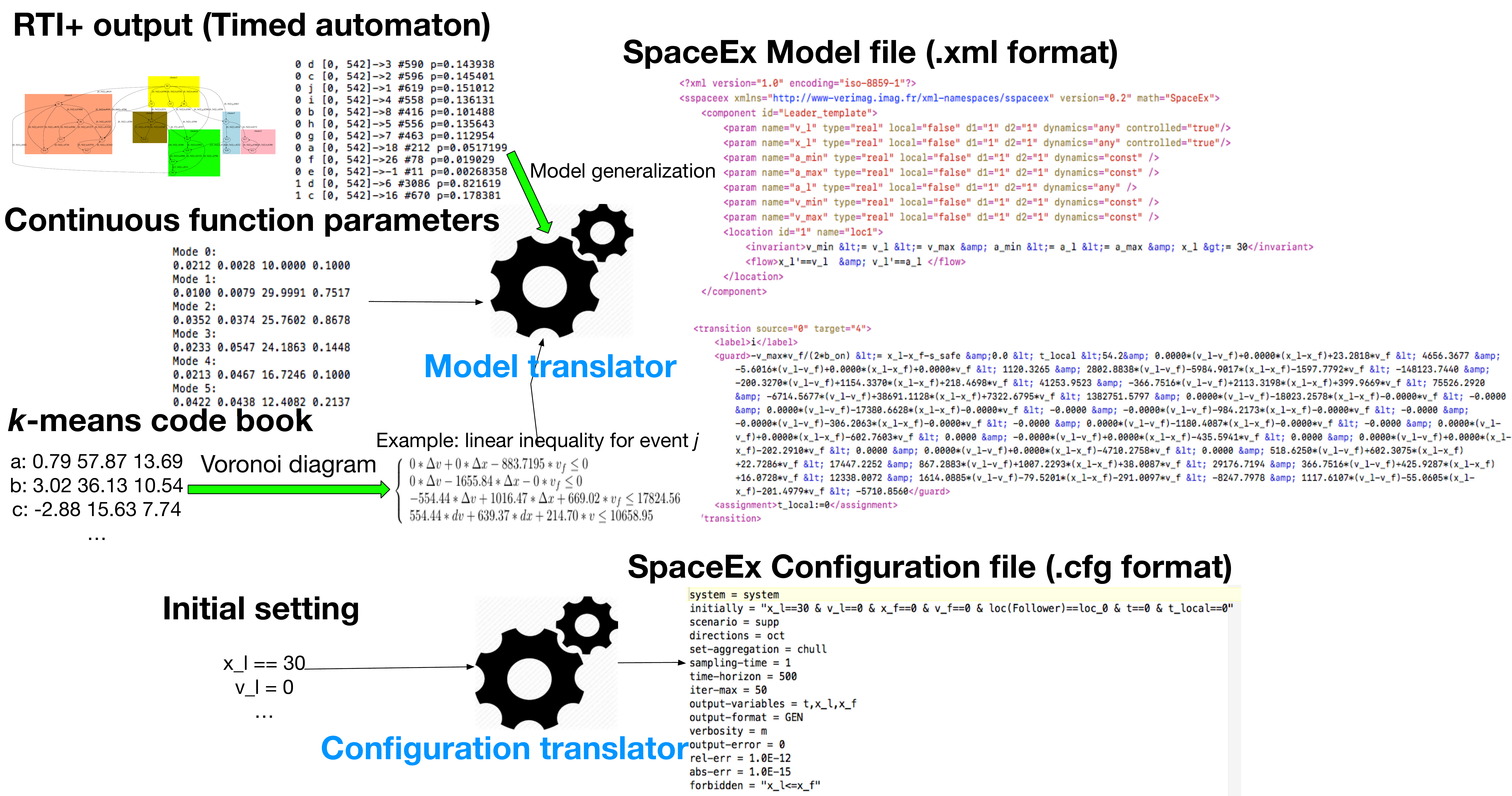}
\caption{Translator MO2SX. The files on the left side is from MOHA and the initial setting. The files on the right side are supported for model checking in SpaceEx.\label{fig:translator}}
\end{figure}

Though SpaceEx is becoming a user-friendly tool, the modeling is still manual. If the model under verification is complex, an automated modelling tool is needed to bypass the tedious modeling process. In our case, we intend to verify a MOHA model, consisting of a timed automaton model, parameters of continuous models in modes, and a discretization of continuous signals into discrete symbols. The translator developed in this paper, \emph{MO2SX}, fills the gap between MOHA and SpaceEX. Users only need to work on learning and tuning parameters of MOHA, and the output model is automatically translated to SpaceEX for safety verification. The input and output files of \emph{MO2SX} are illustrated in Fig. \ref{fig:translator}. MO2SX automatically obtains a SpaceEx model file with 1500 lines of code, which is burdensome for manual modeling.

Guard linearization and model completing are two critical problems we need to address in the translating procedure, and are elaborated below.
\subsubsection{Guard linearization}
\begin{figure}[t]
\centering
\includegraphics[width=0.35\textwidth]{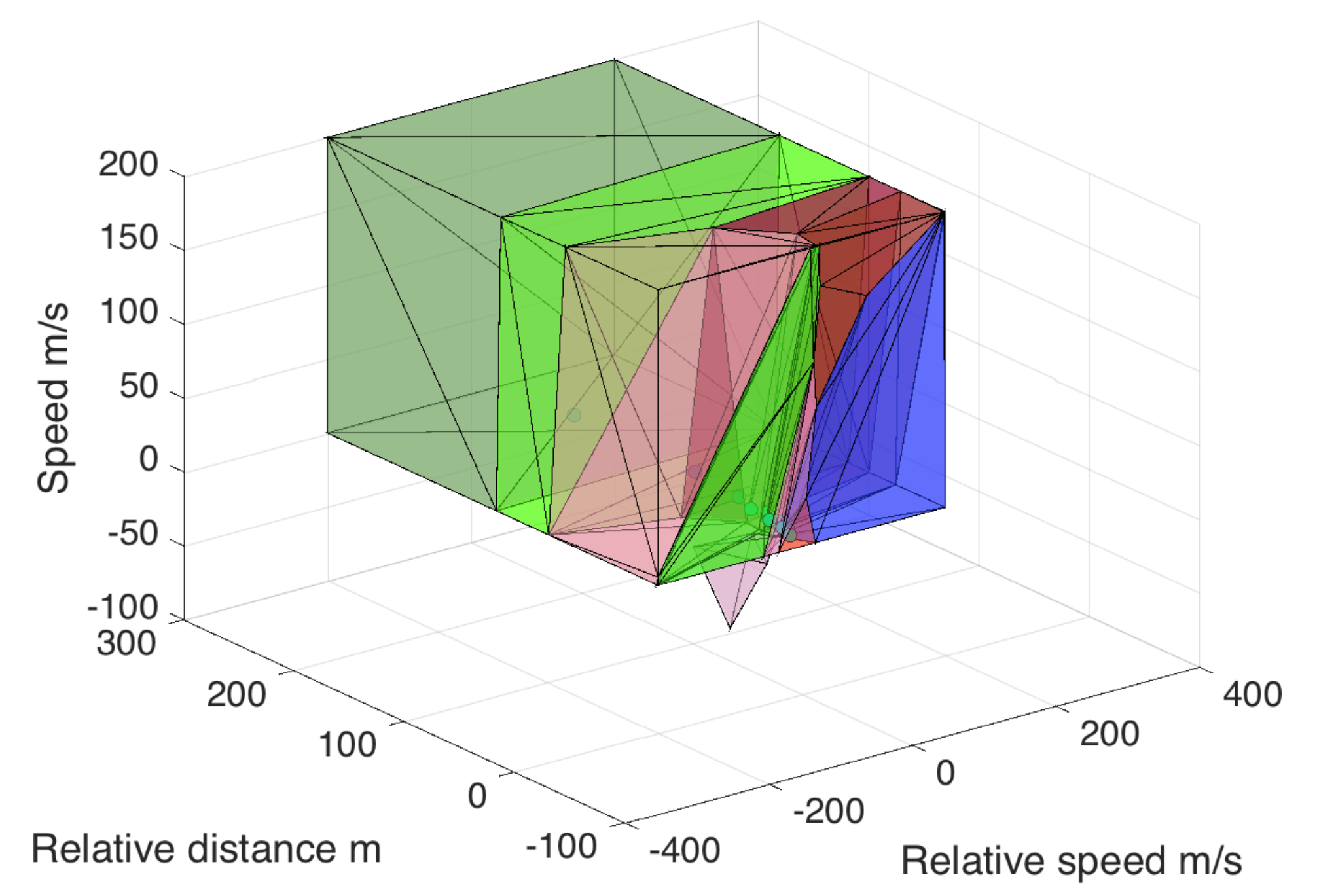}
\caption{Polyhedra obtained by Voronoi diagram linearization. Discrete events are illustrated by different colors. \label{fig:hull}}
\end{figure}

In MOHA, the numeric environmental input is discretized into discrete event symbols according to the closest centroids in the 2-norm, i.e., $S_i = \{x_p:||x_p-m_i||^2 \leq ||x_p-m_j||^2, \forall j, 1\leq j \leq k\}$, where $S_i$ is the assigned index of the centroid (symbol), $x_p$ the numeric data, $m_i, m_j$ centroids, and $k$ the number of centroids. The centroids are learned using the $k$-means clustering algorithm and used to trigger state transitions. This representation is non-linear and not supported by existing hybrid model checkers. To circumvent this issue, we translate the clusters to a bounded three-dimensional Voronoi Diagram \cite{aurenhammer1991}. The main idea is to partition a bounded 3-d space into regions (polyhedra, the number of which is equal to the number of centroids), that are represented by linear inequalities. In each solid polyhedron, all points are closest to its the polyhedron's centroid. Each polyhedron is consists of several hyperplanes, i.e., a conjunction of linear inequalities, as illustrated in Fig. \ref{fig:hull}. Note that MOHA shown in \cite{lin2018} has 10 discrete events from ``a" to ``j", which are essentially symbolic representations from $k$-means clustering on continuous data. Therefore, 10 polyhedra are obtained by the Voronoi diagram. Note that such a linearization is an exact transformation without any approximation error.

\subsubsection{Model Generalization}
\begin{figure}[t]
\centering
\includegraphics[width=0.25\textwidth]{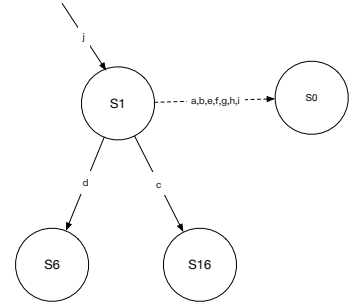}
\caption{An illustrative example of completing outgoing transitions in S1 of MOHA.\label{fig:complete}}
\end{figure}

Due to the limited traffic scenarios in the training data, the learned automaton model is incomplete and does not contain a transition for every possible situation. We complete the model by adding transitions for unsee events and directing them to the initial state. 
Taking $S1$ as an example as shown in Fig. \ref{fig:complete}, the added symbols are the neighboring polyhedra of existing events ``d" and ``c". We obtain these by searching for adjacent polyhedra as illustrated in Fig \ref{fig:hull}. We only require neighboring polyhedra because we assume that trajectories cannot jump between nonadjacent polyhedra (essentially skipping an event). We redirect new transitions to the initial state because this implements a 
type of recovering behavior: when the controller has no idea about what to do next (something unexpected occurs), it makes no assumptions about the past (by returning to the initial state), and assumes any future is possible.

\section{Modeling and experiments}\label{sec:S5}
\begin{figure}[t]
\centering
\includegraphics[width=0.4\textwidth]{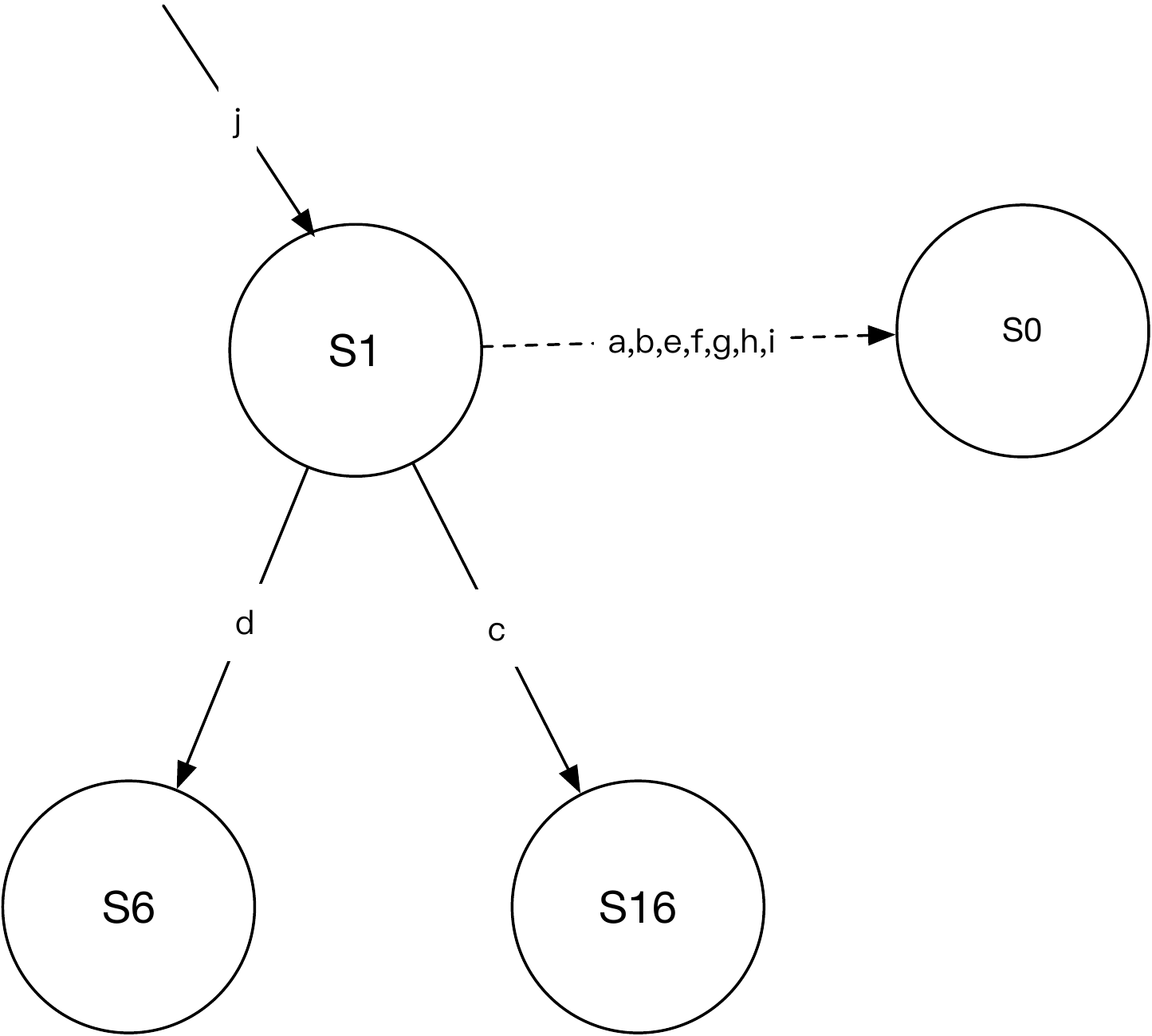}
\caption{Modelling overview of the experiments. \label{fig:model}}
\end{figure}

Our experimental framework (shown in Fig. \ref{fig:model}) consists of two components running in parallel: a nondeterministic \emph{leading vehicle} with constraints about speed and acceleration and a \emph{following vehicle} equipped with a cruise controller. The \emph{autobrake} state is used for handling automatic brake scenarios when the relative distance is small. We will compare the safety performances with and without this state. In this paper, the leading vehicles running in highway and urban traffic are studied:

\begin{itemize}
\item Highway: We adopt the general legitimate range on the highway: 80-120 km/h (see all settings shown in Tab. \ref{tab:normal_speed}). The leading vehicle runs nondeterministically, which means the leading vehicle is not governed by any controller. The free running of the leading vehicle's is only subject to rough physical constraints, e.g., minimum and maximum bounds of speed, acceleration, etc. The speed range is the working condition of a standard ACC system \cite{nissan}.
\item Urban: We adopt the general legitimate range in cities: 10-80 km/h (see all settings shown in Tab. \ref{tab:normal_speed}). The leading vehicle runs nondeterministically. Such a new scenario is for testing the generalization of the model, because the training data of MOHA are from highway traffic.
\end{itemize}

\begin{table}[t]
\centering
\begin{footnotesize}
\caption{Parameter settings in highway scenarios (top) and urban scenarios (bottom) \label{tab:normal_speed}}
\begin{tabular}{lccc}
\toprule
Parameters & values& Parameters & values\\
\midrule
$v_{l\_min}$ (m/s) & 22 & $v_{l\_max}$ (m/s) & 33\\
$v_{f\_min}$ (m/s) & 0 & $v_{f\_max}$ (m/s) & 33\\
 $x_{l0}$ (m) & 150 & $ v_{l0}$ (m/s) & [22,33]\\
$a_{f\_max}$ (m/s\textsuperscript{2}) & 6  & $a_{f\_min}$, $a_{l\_min}$ (m/s\textsuperscript{2})  & -4 \\
$a_{l\_max}$ (m/s\textsuperscript{2}) & 0  &   $v_{f0}$ (m/s) & [22,33] \\
\midrule
\midrule
$v_{l\_min}$ (m/s) & 3 & $v_{l\_max}$ (m/s) & 22\\
$v_{f\_min}$ (m/s) & 0 & $v_{f\_max}$ (m/s) & 22\\
 $x_{l0}$ (m) & 150 & $ v_{l0}$ (m/s) & [3,22]\\
$a_{f\_max}$ (m/s\textsuperscript{2}) & 6  & $a_{f\_min}$, $a_{l\_min}$ (m/s\textsuperscript{2}) & -4 \\
$a_{l\_max}$ (m/s\textsuperscript{2}) & 0  & $ v_{f0}$ (m/s)  & [3,22] \\
\bottomrule
\end{tabular}
\end{footnotesize}
\end{table}

We evaluate three different control strategies:
\begin{itemize}
\item Pure MOHA \textbf{(P-MOHA)}:
A MOHA purely controls the following vehicle without an additional emergency brake state. We will investigate if the Pure-MOHA learned from human car-following behaviors is already safe for cruise control. MOHA with single mode and multiple modes is called \textbf{S-MOHA} and \textbf{M-MOHA} for short, respectively.
\item Autobrake state on basis of braking distance+MOHA \textbf{(BD-MOHA)}: In existing ACCs, a warning notifies the driver to take over or (semi-)automatically switches to a braking state when the relative distance is too short. In this work, a safety state is added to the data-driven P-MOHA to deal with emergency and automatic braking scenarios. The trigger condition of the braking state is that the relative distance $\Delta x$ is smaller than the braking distance $\frac{v \cdot v_{max}}{2*a_{min}}$. Note that theoretically the braking distance is $\frac{v_f^2}{2*a_{min}}$. Due to the limited support functionality of linear equations of SpaceEx, the simplified condition is used instead.
\item Autobrake state on basis of headway-in-time+MOHA \textbf{(HIT-MOHA)}: 
The headway-in-time (HIT) is usually suggested in daily highway driving scenarios.
The follower's desired distance is set to $v_f\times t_{headway}$ for a given $t_{headway}$, i.e., the relative distance should be greater the distance the follower would travel in $t_{headway}$ without reducing speed.
\end{itemize} 

Another motivation for setting an autobrake state is from the theoretical analysis of the minimum deceleration in the Helly model. We take the single mode identified from the natural data with $C1=0.0425$, $C2=0.0051$, $\alpha=22.37$, and $\beta=0.1$. In the worst case, $\Delta v=-33$ m/s, $v_f = 33$ m/s. The full deceleration derived from Equation \ref{equ:helly1} and Equation \ref{equ:helly2} is $-1.68$ m/s\textsuperscript{2}, which is significantly less powerful than the the full deceleration $-4$ m/s\textsuperscript{2} used in this paper.

MOHA is compared with two baseline models in this paper. The first one is a \emph{random follower}. A random follower with nondeterministic dynamics is an over-approximation over any controller. The proportional--integral--derivative (PID) controller is commonly used in existing ACC systems \cite{magdici2017}. Due to the limited functionalities of SpaceEx, the model checker does not allow access to long-term historical variables which are needed for the derivative part of PID. Instead, we use an auxiliary automaton as a one-step-past memory storage, and the \emph{PD controller} is implemented and serves as the second baseline with the form:

\begin{equation}
\left\{
\begin{array}{lr}
d_{des}(i) = d_{safe}+v_f(i)  \\
err = dx(i)-d_{des}(i) \\
a_{pid}(i) = k_p*err(i)+k_d*(err(i)-err(i-1))
\end{array}
\right.
\end{equation}
The parameters are well-tuned on the NGSIM dataset as $K_p=0.8$, $K_d=0.03$, $d_{safe}=20 m$ \cite{lin2018}.

The parameters of vehicle dynamics are also presented in Tab. \ref{tab:normal_speed}. These settings are used in the literature \cite{zhang2018lane}. In both cases, the following vehicle starts tracking at the maximum relative distance detectable by the ACC radar system, i.e., 150 m. The initial states in both cases are uncertain but bounded by reasonable intervals.

\begin{figure*}
        \begin{subfigure}[b]{0.33\textwidth}
                \centering       \includegraphics[width=1.0\linewidth]{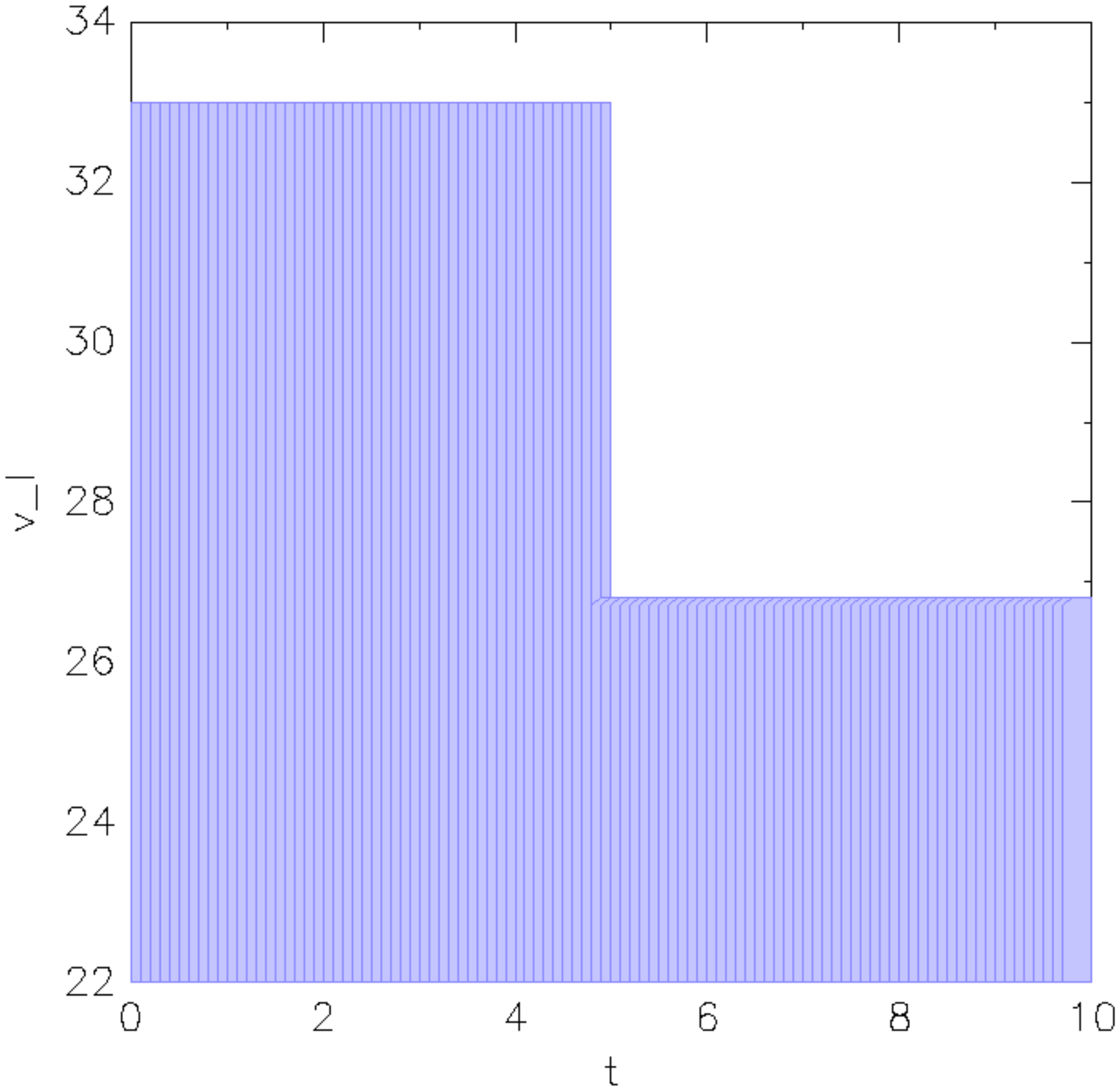}
                \caption{Reachable states of $x_l$ (m) v.s. t (s)}
                \label{fig:xl}
        \end{subfigure}%
        \begin{subfigure}[b]{0.33\textwidth}
                \centering       \includegraphics[width=1.0\linewidth]{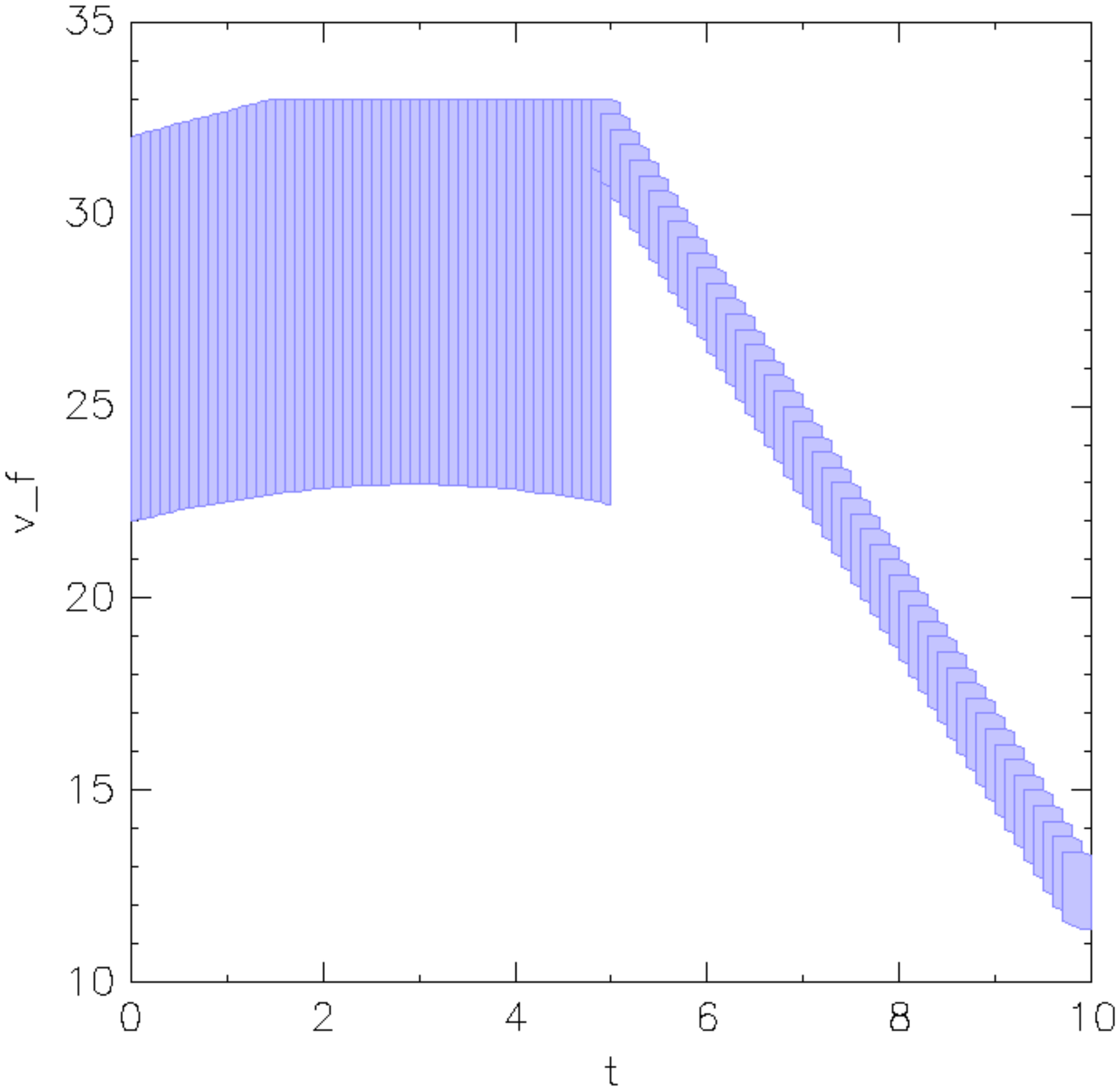}
                \caption{Reachable states of $x_f$ (m) v.s. t (s)}
                \label{fig:xf}
        \end{subfigure}%
        \begin{subfigure}[b]{0.33\textwidth}
                \centering                \includegraphics[width=1.0\linewidth]{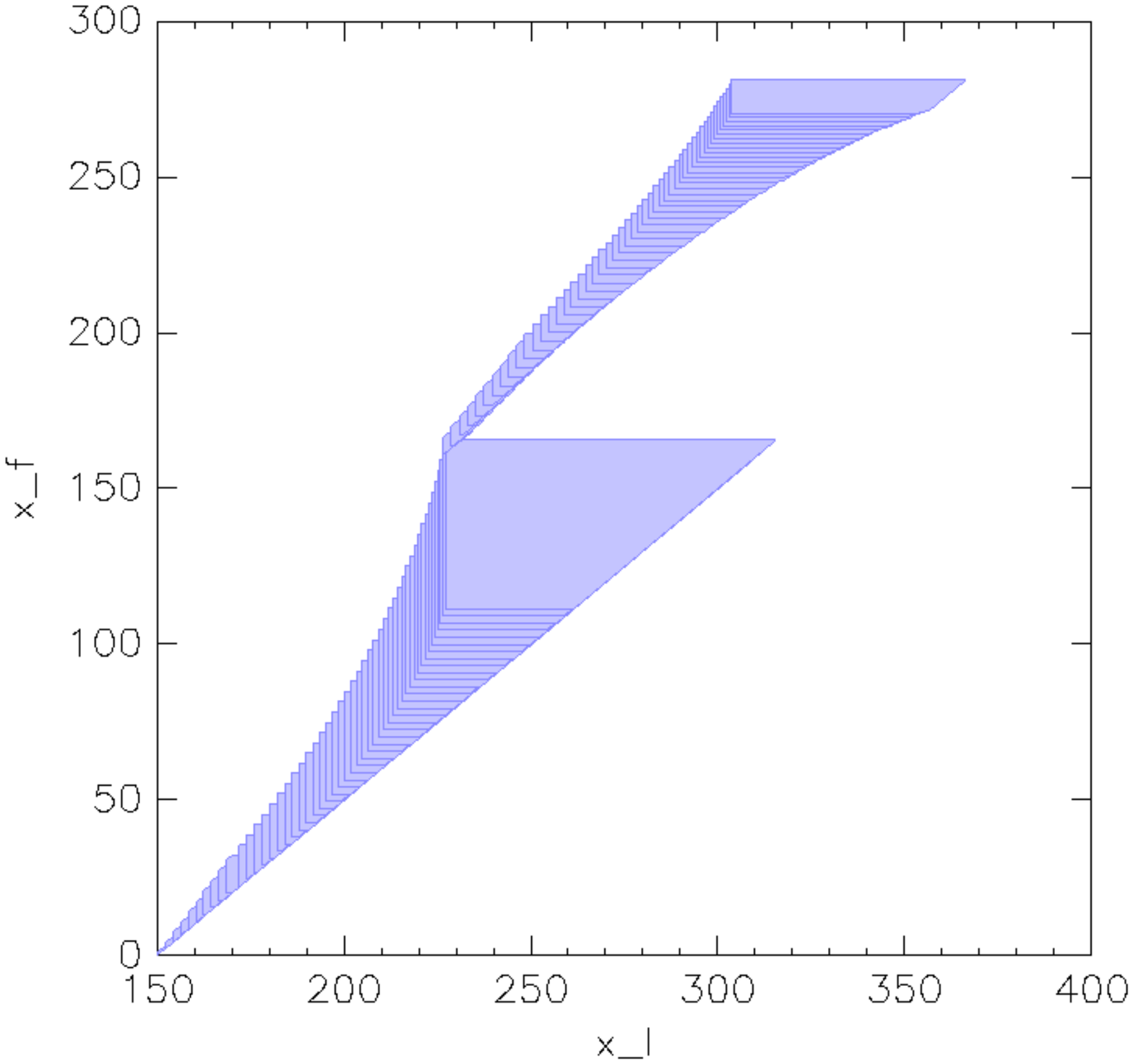}
                \caption{Reachable states $x_l$ (m) vs. $x_f$ (m)}
                \label{fig:xlxf}
        \end{subfigure}%
        \caption{Reachable states of single mode HIT-MOHA in the highway scenario. $x_l$ and $x_f$ are position variables for the leading vehicle and the following vehicle. It can be observed that at around 5 second, the autobrake state is triggered (see the linearly deceleration in subfig (b). After 7 second, the relative speed $v_l-v_f>0$), collision is not possible. The model checker verifies that at any state $x_l>x_f$ (cf. subfig (c)).  }\label{fig:single_moha_reach}
\end{figure*}

An example of the reachability results in the highway scenarios of the single mode HIT-MOHA is shown in Fig. \ref{fig:single_moha_reach}. Tab. \ref{tab:safety_summary} summarizes the safety for all  models and control strategies. It can be observed that the safety state boosts the safety of the controllers. The pure MOHA is not guaranteed to be safe unfortunately.

However, introducing the extra safety state potentially sacrifices the similarity to human car-following behavior. The imitation accuracy, or less formally \emph{human likeness}, is evaluated using a test set from the NGSIM dataset. The main idea is that for each car-following episode, the trajectory of the leading vehicle and the initial status of the following vehicle are provided. The complete trajectory of the following vehicle is generated using controllers and compared with the human drivers' trajectories present in the testing data. A small trajectory difference indicates a better human-likeness score. The results are presented in Tab. \ref{tab:human_likeness}. The score is the mean square error between simulated trajectories and those of human drivers. A feed-forward neural network (FNN) is additionally compared as a baseline of imitation learning with default settings \cite{simonelli2009human,wang2018capturing}. Note that generating whole trajectories is essentially an iterative procedure, i.e., the trajectory at $t+1$ relies on the result at $t$. An additional one-step prediction is shown in Tab. \ref{tab:human_likeness_1step} to demonstrate the actual predictive performance of the learned models. The difference between the results in Tab. \ref{tab:human_likeness} and Tab. \ref{tab:human_likeness_1step} can be seen as the difference between multi-step prediction and one-step prediction.

 \begin{table}[t]
 \begin{center}
 \caption{Safety summary of all models. \label{tab:safety_summary}}
 \centering
 \begin{tabular}{c|c|c|c}
   	\hline
   Scenarios & Model & Condition & Safe?\\
   \hline
\multirow{6}{*}{Highway} & 

P-MOHA & - & ${\times}$\\
& S-MOHA & HIT & ${\surd}$\\
   & M-MOHA & HIT & $\surd$\\
   & Random & HIT & $\surd$\\
   & PD & HIT & ${\surd}$\\
    \cline{2-4}
      & All above & BD & $\surd$\\
   \hline
  
\multirow{6}{*}{Urban} & P-MOHA & - & ${\times}$ \\
& S-MOHA & HIT & ${\surd}$  \\
& M-MOHA & HIT & $\surd$ \\
   & Random & HIT & $\surd$ \\
   & PD & HIT & ${\surd}$ \\
   \cline{2-4}
   & All above & BD & $\surd$ \\
   \hline
 \end{tabular}
 \end{center}
 \end{table}

\begin{table}[h]
\centering
\caption{Human likeness score comparison-multi steps \label{tab:human_likeness}}
{\footnotesize
\begin{tabular}{cccc}
\toprule
& Model & Error (m/s) & Jerk (m/s\textsuperscript{3})\\
\midrule
\multirow{3}{*}{Without safety state} & M-MOHA & \textbf{0.1083} & 0.0037\\
& S-MOHA & 0.1124 & \textbf{0.0029}\\
& PD & 0.1387 & 0.0438\\
& FNN & 0.3451 & 0.0047\\
& Human & - & 0.0574\\
\midrule
\multirow{3}{*}{With safety state} & M-MOHA & \textbf{0.1037}  & 0.0373 \\
& S-MOHA & 0.1089 & \textbf{0.0323}\\
& PD & 0.1391 & 0.0380 \\
& FNN & 0.2411 & 0.0359 \\
& Human & - & 0.0574\\
\bottomrule
\end{tabular}
}
\end{table}

\begin{table}[h]
\centering
\caption{Human likeness score comparison-one step \label{tab:human_likeness_1step}}
{\footnotesize
\begin{tabular}{cccc}
\toprule
& Model & Error (m/s) & jerk (m/s\textsuperscript{3})\\
\midrule
\multirow{3}{*}{Without safety state} & M-MOHA & \textbf{0.0316} & 0.0033\\
& S-MOHA & 0.0317 & \textbf{0.0025}\\
& PD & 0.0543 & 0.0336\\
& FNN & 0.0408 & 0.0048\\
\midrule
\multirow{3}{*}{With safety state} & M-MOHA & \textbf{0.0329}  & 0.0199 \\
& S-MOHA & \textbf{0.0329} & \textbf{0.0195}\\
& PD & 0.0488 & 0.0395 \\
& FNN & 0.0423  & 0.0469 \\
\bottomrule
\end{tabular}
}
\end{table}

From the results, we make the following observations:
\begin{enumerate}
\item The safety is not guaranteed when learning a Pure-MOHA controller. This makes sense because the training data do not contain (near) collisions.
\item Switching to an autobrake state boosts the safety of ACC systems such as MOHA. Among all control strategies, the headway control (HIT) is sufficient and is suggested by us for normal driving scenarios owing to its superior balance between safety and human likeness.
\item The BD is the most conservative control strategy, even though it guarantees full-speed-range scenarios. It is not recommended because the significant large desired relative distance leads to poor car-following performance and traffic jams.
\item Though introducing the safety state slightly deteriorates the car-following performance in one-step prediction, the general performance in whole trajectory control is not jeopardized.
\item MOHA outperforms both the PD and the FNN baselines on human likeness, even when it includes a safety state. There is a significant jump in terms of jerk (sudden braking) when the safety state is triggered.
\end{enumerate}

\section{Conclusion}\label{sec:S6}
In this paper, a framework to automatically learn and verify a hybrid automaton-based adaptive cruise controller is proposed.
The framework consists of a learning-component MOHA and a translator \emph{MO2SX}. MOHA shows a superior performance to human-like car-following, while MO2SX automatically translates MOHA for verification by the SpaceEx hybrid model checker. 
We demonstrate that a MOHA model learned purely from human driving data is not guaranteed to be safe (collision-free) due to the lack of emergency brake scenarios in training data.
Introducing an additional safety state guarantees this safety while maintaining good human likeness scores. To the best of our knowledge, we present the first formally verified cruise control system that is learned from data.

In the near future, we will investigate more driving behaviors learning and verification, e.g., steering control. Another interesting research line is using the model checker as an oracle providing unsafe counterexamples to improve the model learning part.

\addtolength{\textheight}{-12cm}
\bibliography{ref}
\bibliographystyle{unsrt}
\end{document}